\documentclass{article}

\usepackage{arxiv}
\usepackage{multicol}

\usepackage[utf8]{inputenc} 
\usepackage[T1]{fontenc}   
\usepackage{lmodern}
\usepackage[colorlinks=true,citecolor=blue]{hyperref}       
\usepackage{url}            
\usepackage{booktabs}       
\usepackage{amsfonts}       
\usepackage{nicefrac}       
\usepackage{microtype}      
\usepackage{lipsum}		
\usepackage{graphicx}
\usepackage{natbib}
\usepackage{dirtytalk}

\usepackage{doi}
\usepackage{multirow}
\usepackage{booktabs}
\usepackage{xcolor}

\usepackage{adjustbox}
\newcommand{\hlc}[2]{\adjustbox{cframe=#1 1pt}{#2}}

\definecolor{Persistent}{rgb}{1.0, 0.94, 0.0}
\definecolor{Comparison}{rgb}{0.86, 0.82, 1.0}
\definecolor{Prior}{rgb}{1.0, 0.71, 0.76}
\definecolor{Changes}{rgb}{0.53, 0.81, 0.98}
\definecolor{Recurrence}{rgb}{0.69, 0.19, 0.38}
\definecolor{Unchanged}{rgb}{0.6, 0.73, 0.45}
\definecolor{Again}{rgb}{0.93, 0.53, 0.18}

\definecolor{Increased}{rgb}{0.5, 0.3, 0.1}
\definecolor{Stable}{rgb}{0.2, 0.5, 0.1}
\definecolor{Decreased}{rgb}{0.1, 0.1, 0.9}
\definecolor{Previous}{rgb}{0.9, 0.1, 0.9}

\definecolor{Worse}{RGB}{99, 114, 242}
\definecolor{Change}{RGB}{91, 200, 154}
\definecolor{Unchanged}{RGB}{222, 95, 70}
\definecolor{BLACK}{RGB}{0, 0, 0}

\title{\Large{Improving Radiology Report Generation Systems by Removing Hallucinated References to Non-existent Priors}}

\author{ Vignav Ramesh\thanks{These authors contributed equally} \\
	Harvard University \\
	\texttt{vignavramesh@college.harvard.edu} \\
	\And
	Nathan A. Chi$^{*}$ \\
	Stanford University \\
	\texttt{nchi1@stanford.edu} \\
	\AND
	Pranav Rajpurkar \\
	Harvard Medical School \\
	\texttt{pranav\_rajpurkar@hms.harvard.edu} \\
}

\date{}

\renewcommand{\shorttitle}{\textit{Removing References to Priors}}

\hypersetup{
pdftitle={A template for the arxiv style},
pdfsubject={q-bio.NC, q-bio.QM},
pdfauthor={David S.~Hippocampus, Elias D.~Striatum},
pdfkeywords={First keyword, Second keyword, More},
}

\begin{document}

\maketitle

\begin{multicols}{2}

\begin{abstract}
	Current deep learning models trained to generate radiology reports from chest radiographs are capable of producing clinically accurate, clear, and actionable text that can advance patient care. However, such systems all succumb to the same problem: making hallucinated references to non-existent prior reports. Such hallucinations occur because these models are trained on datasets of real-world patient reports that inherently refer to priors. To this end, we propose two methods to remove references to priors in radiology reports: (1) a GPT-3-based few-shot approach to rewrite medical reports without references to priors; and (2) a BioBERT-based token classification approach to directly remove words referring to priors. We use the aforementioned approaches to modify MIMIC-CXR, a publicly available dataset of chest X-rays and their associated free-text radiology reports; we then retrain CXR-RePaiR, a radiology report generation system, on the adapted MIMIC-CXR dataset. We find that our re-trained model---which we call CXR-ReDonE---outperforms previous report generation methods on clinical metrics, achieving an average \textsc{BERTScore} of 0.2351 ($2.57\%$ absolute improvement). We expect our approach to be broadly valuable in enabling current radiology report generation systems to be more directly integrated into clinical pipelines. Our code, data, and pre-trained model weights are made available at \href{https://github.com/rajpurkarlab/CXR-ReDonE}{this link}.

\end{abstract}

\keywords{free-text radiology reports \and references to priors \and generation \and retrieval \and large language models
}

\section{Introduction}
\label{sec:intro}

Writing radiology reports is a tedious and labor-intensive process, requiring trained specialists to conduct in-depth analyses of chest radiographs and create detailed reports of their findings. This process is also inherently restricted by a variety of human limitations, including the experience of the radiologist and availability of medical support staff. Therefore, automatically generating free-text radiology reports from chest radiographs has immense clinical value.

Current approaches to generate radiology reports from chest X-rays (CXR-RePaiR, R2Gen, $\mathcal{M}^2$ Trans, etc.) have achieved relative success in producing complete, consistent, and clinically accurate reports \citep{pmlr-v158-endo21a, chen2020generating, miura2021improving, johnson2019mimic, 9757742}. Nevertheless, these models each have a key limitation: since they are trained on datasets of real-world reports (MIMIC-CXR, Indiana University Chest X-ray Collection, etc.) which refer to prior reports, their outputted reports often contain references to non-existent priors \citep{Johnson2019, indiana}. 


\begin{figure*}[!ht]
    \centering
    \includegraphics[width=\textwidth]{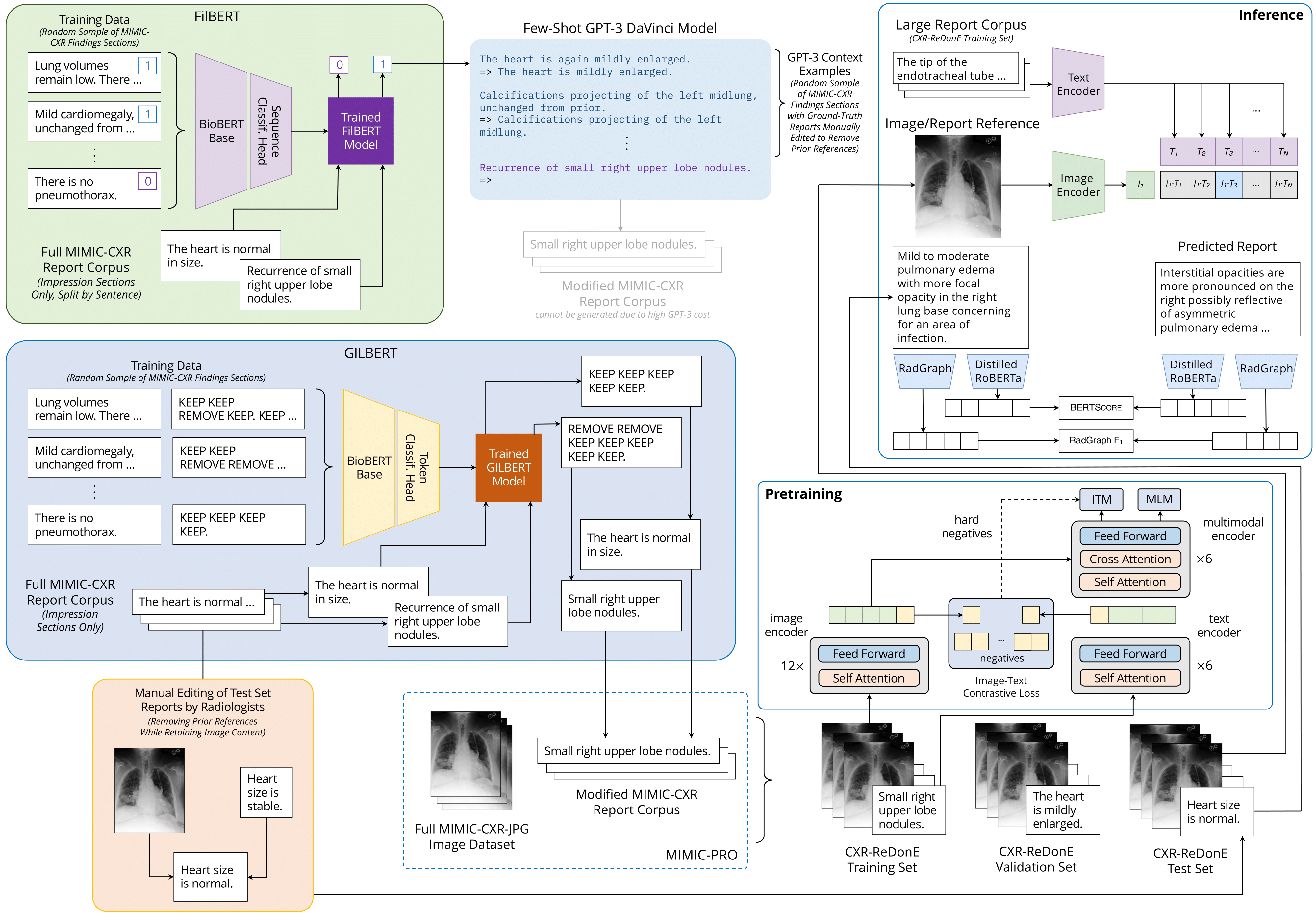} 
    \caption{\textbf{CXR-ReDonE pipeline.} We first generate MIMIC-PRO by passing reports from MIMIC-CXR through GILBERT. It should be noted that we also investigate a secondary pathway to synthesize MIMIC-PRO---the two-step pipeline FilBERT+GPT-3---but do not employ it due to its decreased accuracy and higher cost. We then train CXR-ReDonE by passing reports and chest X-rays from MIMIC-PRO through a text encoder and image encoder, respectively. Finally, CXR-ReDonE outputs the generated report with the highest dot-product similarity between the text and image embeddings, and performance metrics are calculated by comparing the ground truth to the predicted reports.}
    \label{fig:figure1}
\end{figure*}

To address this issue, we propose \textit{\underline{C}ontrastive \underline{X}-\underline{R}ay \underline{Re}port \underline{D}eterminati\underline{on} \underline{E}mploying Prior Reference Removal} (CXR-ReDonE), an improved radiology report generation approach that eliminates nearly all hallucinated references to priors (Figure 1). CXR-ReDonE’s unique contribution lies in its 
novel data preprocessing step: it is trained on MIMIC-PRO, our adaptation of \textit{\underline{MIMIC}-CXR with \underline{P}rior \underline{R}eferences \underline{O}mitted}. 

Specifically, we investigate two separate approaches to generate MIMIC-PRO: \textbf{(1)} FilBERT+GPT-3, a two-step pipeline to rewrite entire medical reports with prior references removed; and \textbf{(2)} GILBERT, a BioBERT model fine-tuned to remove references to priors at the token level \citep{brown2020language, DBLP:journals/corr/abs-1901-08746}. While both approaches are viable, we employ GILBERT to create MIMIC-PRO due to its higher performance, lower cost, and increased scalability.

CXR-ReDonE outperforms existing state-of-the-art report generation methods, achieving an average \textsc{BERTScore} \citep{zhang2020bertscore} of 0.2351 ($2.57\%$ absolute improvement over the highest-performing baseline). We anticipate that our approach will improve performance of current supervised radiology report generation systems---beyond just CXR-RePaiR---on clinical metrics. Given that generated reports without prior references are more aligned with radiologist-created reports, we find that our approach can be broadly valuable in enabling such systems to be more directly integrated into clinical workflows.

\begin{table*}[!htbp]
\scriptsize
  {\caption{A sample report from MIMIC-CXR, as well as examples of different methods' generated reports. References to priors---true for MIMIC-CXR and hallucinated for CXR-RePaiR, R2Gen, and $\mathcal{M}^2$ Trans---are boxed.\\}}
  {\begin{tabular}{p{0.355\linewidth} p{0.18\linewidth} p{0.18\linewidth} p{0.18\linewidth}}
  \toprule
  \textbf{MIMIC-CXR} & \textbf{CXR-RePaiR} & \textbf{R2Gen} & \textbf{$\mathcal{M}^2$ Trans} \\
  \midrule
  Frontal and lateral radiographs of the chest demonstrate fractured lower three median sternotomy wires \hlc{BLACK}{which are unchanged from \_\_\_}. \hlc{BLACK}{Compared to the prior radiograph}, there is \hlc{BLACK}{increase} in airspace opacity at the right lung base, consistent with pneumonia. The remainder of the lungs is \hlc{BLACK}{unchanged from the prior radiograph}. The cardiac contour is slightly enlarged, \hlc{BLACK}{unchanged from the prior radiograph}. No pleural effusion or pneumothorax is seen.
  & 
  
  {Cardiomegaly} with left {pleural} {effusion} and {mild} {edema}. The severe {enlargement} {of cardiac 
  silhouette} \hlc{BLACK}{may have progressed} the volume of presumed right {pleural effusion} is impossible to assess on a single frontal view.
  &
  There are relatively low lung volumes. There is \hlc{BLACK}{diffuse increase} in interstitial markings bilaterally which may be due to mild interstitial {edema} versus atypical infection. \hlc{BLACK}{The cardiac and me-} \hlc{BLACK}{diastinal silhouettes} \hlc{BLACK}{are stable}. 
 
  &
  Heart size \hlc{BLACK}{remains} mildly enlarged. \hlc{BLACK}{The mediastinal and} \hlc{BLACK}{hilar contours are} \hlc{BLACK}{unchanged}. There is mild pulmonary edema. There is no focal consolidation. There is no large pleural effusion or pneumothorax.

  \\
    \midrule

  \end{tabular}}
\end{table*}

\section{Related Work}
\label{sec:2}
Previous radiology report generation approaches have made use of an assortment of techniques, including applying novel CNNs, RNNs, and LSTMs; fine-tuning image recognition models and Transformers; and developing larger-scale ensemble methods \citep{monshi2020deep, rubin2018large, Jing_2018, wang2018tienet, alfarghaly2021automated, chen2020generating}.

However, all of these models---regardless of their structure---tend to generate incomplete, incorrect, or hallucinatory statements within reports (Table 1). Even state-of-the-art models still generate reports with missing or false references when evaluated by trained radiologists \citep{alfarghaly2021automated, liu2019clinically}. Such issues stem from those of abstract text generation models in general \citep{maynez2020faithfulness}. \citet{miura2021improving}, for instance, noted that the standard teacher-forcing training algorithm \citep{williams1989learning} employed by several report generation models results in discordance between training and test environments, thereby causing the production of factual hallucinations---referring to non-present conditions while overlooking existing ones.

Hence, more recent works have focused on improving the factual completeness and consistency of radiology report generations. The most popular approach in this regard has been designing evaluation metrics more suited to the domain of radiology reports. Previous report generation models have attained promising results on standard natural language generation (NLG) metrics such as CIDEr \citep{DBLP:journals/corr/VedantamZP14a}, BLEU \citep{papineni-etal-2002-bleu}, and METEOR \citep{banerjee-lavie-2005-meteor}; however, such metrics tend to reward superficial textual similarities (i.e., same diction, phrase structure, word order, etc.) rather than semantic or ontological similarities---that is, similarities in the actual diagnosis conveyed by the report \citep{boag-baselines, chen-2020, miura2021improving}. 

To address this issue, a variety of metrics have been posed, both to reward clinical accuracy during training, and to guide model evaluation. For instance, \citet{miura2021improving} proposed Exact Entity Match Reward ($\mathrm{fact}_{\mathrm{ENT}}$) to measure generated reports' coverage of radiological entities along with Entailing Entity Match Reward ($\mathrm{fact}_{\mathrm{ENTNLI}}$), which rewards inferential consistency by extending $\mathrm{fact}_{\mathrm{ENT}}$ with a natural language inference (NLI) model; they then optimize these rewards using a reinforcement learning (RL) model. Similarly, \citet{zhang2020bertscore} introduced the \textsc{BERTScore} metric for evaluation that computes token similarity using contextual embeddings, thereby capturing semantic similarity (diagnostic similarity in the radiology report domain). Other models aiming to improve the quality of generated radiology reports have employed question answering (QA) \citep{wang-etal-2020-asking} and content matching constraint \citep{wang-etal-2020-towards} approaches, among other techniques; however, such models lack either sufficient performance or generalization capabilities.

Despite these efforts, none of the existing approaches to improve radiology report generation systems' clinical accuracy address the issue of references to non-existent priors---which remain a critical obstacle in generating factually complete and consistent reports. As such, we propose two prior reference removal approaches, FilBERT+GPT-3 and GILBERT, to eliminate these hallucinations. We provide a proof of concept of these models' abilities in the form of retraining CXR-RePaiR, an existing Transformer-based report generation model, on MIMIC-PRO \citep{endo2021retrieval}. Rather than structuring the task of report creation as one of text generation, CXR-RePaiR adopts a retrieval approach, which enables it to benefit from the limited space of possible findings and diagnoses in chest radiograph-associated radiology reports. We expect that our proposed methodology has the capacity to improve clinical performance of all existing radiology report generation systems.

\section{Data and Implementation}
The entirety of our data comes from MIMIC-CXR, a publicly available dataset of chest X-ray images and associated free-text radiology reports, which constitutes 377,110 images taken from 227,835 radiological studies. We make use of a curated set of 226,759 reports (henceforth, MIMIC-CXR will refer to this curated set). CXR-ReDonE is trained on MIMIC-PRO (created by running GILBERT on all elements of the MIMIC-CXR report corpus) and evaluated on an independent expert-edited evaluation set from MIMIC-CXR (Section \ref{sec:33}).

\subsection{Data Exploration}
\label{sec:31}
We begin by examining the space of references to prior reports, given that the possibilities for prior reference expression in radiology reports are finite. Through manual exploration of the data, we determine that prior references in MIMIC-CXR are grammatical variations on any of 18 main keywords (Table 2).

Based on the data, we find that the most common references to priors are by far \textit{change}, \textit{unchanged}, and \textit{prior}. In total, 173,822 reports (76.3\%)  in MIMIC-CXR contain at least one reference to one of the 18 main keywords. That is, a substantial majority of MIMIC-CXR reports make references to priors. It is worth noting that the keyword \textit{change} does not always refer to prior reports, due to a diversity of possible usages. Most notably, when paired with qualifiers like \textit{emphysematous} or \textit{bony}, \textit{change} refers to a qualitative finding rather than a comparative development from a prior condition. To inform our classification of \textit{change} references, we consulted a physician in order to determine which \textit{change} keywords did and did not refer to priors (Table 8).

\subsection{Shared Corpus}
\label{sec:32}
To develop FilBERT and GILBERT, we create a \textit{shared corpus} of pairs of original radiology reports and their reworded versions with removed references to priors. Specifically, we manually curate a series of 103 reports that are representative of the full set of prior keywords, then create ground truth reworded reports by removing references to priors (e.g., \say{No interval change from yesterday. Tubes and lines in adequate position.} $\Rightarrow$ \say{Tubes and lines in adequate position.}). 

We use the shared corpus to train and evaluate FilBERT and GILBERT; specifically, we employ a shuffled 80-20 train-test split. A patient that appears in the train set does not appear in the test set. Additionally, by design, the proportion of references to priors per sentence is much higher in the shared corpus than in the MIMIC-CXR dataset as a whole, so that the full set of keywords can be represented in a limited space.

\begin{table*}[!ht]
    \small
    \centering
    \begin{tabular}{lll}
    \toprule
    {\textbf{Keyword}} & {\textbf{Frequency}} & {\textbf{Relative}} \\
    \midrule
    \textit{Total} & \textit{173822} & \textit{0.763}\\
    Change & 105244 & 0.462\\
    Unchanged & 65037 & 0.285\\
    Prior & 56572 & 0.248\\
    Stable & 43340 & 0.190\\
    Interval & 42124 & 0.185\\
    Previous & 34155 & 0.150\\
    Again & 25257 & 0.111 \\
    Increased & 25163 & 0.110\\
    Improve & 21941 & 0.096\\
    Remain & 20351 & 0.089\\
    Worse & 17197 & 0.075\\
    Persistent & 12371 & 0.054\\
    Removal & 12068 & 0.053\\
    Similar & 11694 & 0.051\\
    Earlier & 11277 & 0.049\\
    Decreased & 9919 & 0.044\\
    Recurrence & 2872 & 0.012\\
    Redemonstrate & 1099 & 0.005\\
    \bottomrule
    \end{tabular}
    
    \caption{Frequencies and relative frequencies of a keyword referring to change appearing in a radiology report.}
\end{table*}

\subsection{Report Generation Evaluation Set}
\label{sec:33}
In order to test the capabilities of our report generation model CXR-ReDonE, we recruit a team of one board-certified radiologist and two fourth year medical students to create a ground truth test set of reports without references to priors. In particular, we make use of the standardized MIMIC-CXR test set containing 2,192 images and associated reports \citep{endo2021retrieval}. We provide the medical annotators with the directive to either remove or rewrite references to priors in the test set reports. For instance, \say{no interval change from prior CT} is a phrase that can be removed completely, while \say{heart size is stable} must be changed to a description of the heart's current state (e.g., \say{heart size is abnormal}) rather than simply removed.

\section{Methods}
\label{sec:methods}
We examine two methods to remove references to priors in radiology reports: (1) FilBERT+GPT-3: \textit{rewriting} report sentences flagged as containing references to priors to remove all such references; and (2) GILBERT: directly \textit{removing} tokens referring to priors using a BioBERT token-level classification model. Table 3 contains an example report modified by FilBERT+GPT-3 and GILBERT.

Given that it is less costly to run, we employ GILBERT to create MIMIC-PRO. We then develop CXR-ReDonE by retraining CXR-RePaiR on MIMIC-PRO for the task of predicting the impression section of a radiology report from a given chest X-ray.

\subsection{FilBERT+GPT-3: A Two-Step Approach to Prior Reference Removal}
We use a GPT-3 DaVinci model to rewrite report sentences with removed references to priors. However, running GPT-3 on every sentence of each report would be prohibitively costly (requiring an estimated \$92,000 to process the entirety of MIMIC-CXR). Therefore, we propose an additional preprocessing step: FilBERT, a sequence classification model that flags individual sentences as containing references to priors. That is, FilBERT effectively \textit{filters} out all sentences without references to priors, allowing GPT-3 to only rewrite sentences that do. By running GPT-3 in conjunction with FilBERT, we significantly reduce wasted computational and financial effort---resulting in a projected total cost of \$16,560 (a more than five-fold reduction).

\subsubsection{FilBERT: \textit{\underline{Fil}tering Sentence-Level References to Priors with Bio\underline{BERT}}}

Given the impracticality of labeling large radiology report corpora, we investigate fine-tuning an existing, domain-specific language model to classify whether or not reports incude references to priors. To this end, we examine BioBERT, a BERT model pre-trained on a variety of biomedical texts, ranging from medical abstracts to full biomedical papers \citep{lee2020biobert}. FilBERT contains a BioBERT base architecture with a sequence classification head that is finetuned on the shared corpus for the task of flagging sentences with prior references.

\subsubsection{GPT-3 for Rewriting Reports}
After identifying the sentences containing references to priors with FilBERT, we feed the flagged data into GPT-3 DaVinci, which generates reworded alternatives for each input sentence. In order to engineer a prompt for GPT-3, we first identified sentences that would be representative of the full list of prior reference keywords and subsequently selected a set of 29 samples as our contextual examples. See Appendix \ref{apd:first} for a full discussion of our hyperparameter search and prompt development process.

\subsection{GILBERT: Generating \underline{I}n-text \underline{L}abels of References to Priors with Bio\underline{BERT}}

\begin{table*}[!ht] 
\centering
\scriptsize

  {\caption{Example of different methods' reports with removed references to priors compared to the ground truth report. Prior references are color coded to improve readability. See Appendix \ref{apd:sec} for additional examples.\\}}
  {\begin{tabular}{p{0.18\linewidth} p{0.18\linewidth} p{0.18\linewidth} p{0.18\linewidth}}
  \toprule
  \textbf{Original} & \textbf{Ground Truth} & \textbf{FilBERT+GPT-3} & \textbf{GILBERT}\\
  \midrule
  
 \hlc{Comparison}{Comparison made to} \hlc{Prior}{prior study from \_\_\_}. 
 
 &

 &
 \hlc{Comparison}{Comparison is made to} \hlc{Prior}{the prior study}.
 &

 \\
 
 There is \hlc{Again}{again} seen moderate congestive heart failure with \hlc{Increased}{increased} vascular cephalization, \hlc{Stable}{stable}.

  &
  There is seen moderate congestive heart failure with vascular cephalization. 
  
  &
Moderate congestive heart failure with \hlc{Increased}{increased} vascular cephalization. 
  
  & 
  There is seen moderate congestive heart failure vascular cephalization. 
  \\
  
  There are large bilateral pleural effusions \hlc{Decreased}{but decreased} \hlc{Previous}{since previous}. There is cardiomegaly. 
  
  & 
  
  There are large bilateral pleural effusions. There is cardiomegaly.
  
  & 
  There are large bilateral pleural effusions. There is cardiomegaly.
  &
  There are large bilateral pleural effusions. There is cardiomegaly. 
  \\

  \bottomrule
\end{tabular}}

  {\begin{tabular}{c c c c c c c c c c}
  \hlc{Comparison}{comparison} & \hlc{Prior}{prior} & \hlc{Recurrence}{recurrence} & \hlc{Again}{again} & \hlc{Increased}{increased} & \hlc{Stable}{stable} & \hlc{Decreased}{decreased} & \hlc{Previous}{previous}\\
  \end{tabular}}\\
\end{table*}

Next, we introduce GILBERT, a BioBERT model tasked with classifying tokens as referring to priors or not. Specifically, GILBERT casts the radiology report labeling process as a named entity recognition (NER) task; the model classifies each token in an inputted report as either REMOVE (denoting that the token constitutes a reference to a prior and should be removed from the outputted report) or KEEP (indicating that the token does not constitute a reference to a prior and should be included in the final report). GILBERT uses a modified version of the shared corpus for training, where the reworded report is replaced with a string of KEEP and REMOVE tokens (e.g., \say{hilar prominence suggestive of pulmonary hypertension, unchanged} $\Rightarrow$ \say{KEEP KEEP KEEP KEEP KEEP KEEP REMOVE}).

\subsubsection{Model Structure}
GILBERT contains a BioBERT base architecture, with a token classification head placed on top to allow for predictions at the token level. As BERT relies on wordpiece tokenization \citep{wuwuwuw}, we modify GILBERT’s accompanying tokenizer to label all wordpiece units in the dataset as either KEEP or REMOVE, rather than just the first subunit of a each word.

GILBERT employs the same training process as FilBERT, except that a token classification rather than sequence classification head is fine-tuned.

\subsection{CXR-ReDonE}

\begin{table*}[!ht]
\centering
\small

  {\caption{Evaluation of CXR-ReDonE method on expert-edited test set, for report-level retrieval and sentence-level retrieval with $k\in\{1,2,3\}$, after training on MIMIC-CXR and MIMIC-PRO. Metrics employed are \textsc{BERTScore}, $s_{emb}$, and RadGraph $F_1$. We find that, irrespective of $k$, our approach outperforms the baseline on \underline{all} clinical metrics. Here, \textit{italics} denote improvement over the baseline, while \textbf{bold} denotes the highest value across the board.\\}}
  {\begin{tabular}{llccc}
  \toprule
  \multirow{2}{*}{\textbf{$k$}} & 
  \multirow{2}{*}{\textbf{Training Dataset}} &
  \multicolumn{3}{c}{\textbf{Evaluation Metrics}} \\
  & & \textsc{\textbf{BERTScore}} & \textbf{$s_{emb}$} & \textbf{RadGraph $F_1$} \\
  \midrule
  N/A (report-level & \multirow{1}{*}{MIMIC-CXR (Baseline)}
   & 0.2083 & 0.3410 & 0.0895 
  
  \\
   retrieval) & \multirow{1}{*}{MIMIC-PRO (Ours)} & \textit{0.2160} & \textit{0.3601} & \textit{0.0925}

\\

\midrule
  \multirow{2}{*}{1} & \multirow{1}{*}{MIMIC-CXR (Baseline)}  & 0.2129 & 0.3880 & 0.0838 
  
  \\
   & \multirow{1}{*}{MIMIC-PRO (Ours)} & \textit{0.2159} & \textbf{\textit{0.3967}} & \textit{0.0864}

\\

\midrule
  \multirow{2}{*}{2} & \multirow{1}{*}{MIMIC-CXR (Baseline)}  & 0.2292 & 0.3822 & 0.1045 
  
  \\
   & \multirow{1}{*}{MIMIC-PRO (Ours)} &\textbf{ \textit{0.2351} }& \textit{0.3859} & \textit{0.1056}
   
\\

\midrule
  \multirow{2}{*}{3} & \multirow{1}{*}{MIMIC-CXR (Baseline)}  & 0.2179 & 0.3710 & 0.1083 
  
  \\
   & \multirow{1}{*}{MIMIC-PRO (Ours)} & \textit{0.2254} & \textit{0.3779} & \textbf{\textit{0.1112}}

\\
  \bottomrule
  
  \end{tabular}}
\end{table*}

After generating MIMIC-PRO with GILBERT, we then develop CXR-ReDonE by retraining CXR-RePaiR, a retrieval-based radiology report system  \citep{endo2021retrieval}. However, we investigate one key architectural change in CXR-ReDonE: replacing the CLIP base model with an ALBEF counterpart, given the latter's higher performance on a variety of vision-language downstream tasks \citep{li2021align}.

As in CXR-RePaiR, the problem of generating radiology reports is structured as a retrieval task from report corpus $\mathcal{R} = \{r_1, ..., r_n\}$. Given a chest X-ray $x$, CXR-ReDonE creates report $\hat p$, which is either report $r \in \mathcal{R}$ or composite report $s$, a combination of $k$ report sentences from the set of all sentences in the retrieval corpus. Note that, for a given report $r$ or sentence $s$, the base model creates a text embedding for $r$ or $s$ and an image embedding for $x$, then calculates similarity score $f(r, x) = g(r) \cdot h(x) = T \cdot I$ or $f(s, x) = g(s) \cdot h(x)$; $\hat p$ is chosen as the report which maximizes this dot product.

\section{Experiments}

\subsection{Evaluation Setup}
To evaluate the performance of FilBERT+GPT-3 and GILBERT, we calculate the $F_1$ score using the output of a difference checker algorithm comparing our original, modified (outputted by our models), and ground truth reports. In the context of our study, true positives denote tokens removed from the original report in both the modified and ground truth reports, false positives denote tokens removed in the modified report but not in the ground truth report, and false negatives denote tokens kept in the modified report but removed in the ground truth report.

\subsection{FilBERT+GPT-3}
On its sentence classification task, FilBERT attains a high accuracy of 0.907 on a held-out test set from the shared corpus (Section \ref{sec:32}). Moreover, the model's error distribution---more false positives than false negatives---is well-adapted to our task. During evaluation, FilBERT misclassifies only 5.55\% of all reports as false positives and 3.70\% of reports as false negatives, which is preferable given that sentences incorrectly labeled as containing references to priors should be unaffected when fed into GPT-3. We empirically determine that a discrimination threshold of 0.5 is optimal; shifting the threshold results in either a larger proportion of false negatives or a lower total accuracy. 

The combined FilBERT+GPT-3 pipeline attains an $F_1$ score of 0.56.

\begin{table*}[!ht]
\centering
\scriptsize

  {\caption{Example of different methods’ generated reports compared to the ground truth report. Report-level retrieval is employed. See Appendix \ref{apd:sec} for additional examples.\\}}
  {\begin{tabular}{p{0.298\linewidth} p{0.298\linewidth} p{0.298\linewidth} }
  \toprule
  \textbf{Ground Truth} & \textbf{Model Trained on MIMIC-CXR} & \textbf{Model Trained on MIMIC-PRO}\\
  \midrule
  
A right IJ catheter terminating at the mid right atrium and multiple sternal wires and mediastinal clips are in position.
 &
 
 Various support and monitoring devices \hlc{Decreased}{removed} \hlc{Previous}{residual} right internal jugular catheter in place and no visible pneumothorax. Cardiomediastinal contours \hlc{Stable}{stable}. Lungs are remarkable for bibasilar patchy atelectasis the left on the right.
 
 &

Placement of right internal jugular central venous catheter terminating at the cavoatrial junction without evidence of pneumothorax. Enteric tube grossly courses below the level of the diaphragm, inferior aspect not included on the image. \\

  \bottomrule
\end{tabular}}

  {\begin{tabular}{c c c}
  \hlc{Stable}{stable} & \hlc{Decreased}{removed} & \hlc{Previous}{residual}\\
  \end{tabular}}\\
  
\end{table*}

\subsection{GILBERT}

GILBERT achieves an $F_1$ score of 0.84 on its corresponding held-out test set. This demonstrates that GILBERT's generated reports are notably closer in semantics to the ground truth reports than FilBERT+GPT-3's. More significantly, a study of all 226,759 reports in the MIMIC-CXR and MIMIC-PRO datasets shows that the number of references to priors decreases drastically, from 259,376 instances of keywords (Section \ref{sec:31}) denoting prior references in MIMIC-CXR to 82,074 in MIMIC-PRO---a $>$$68.3\%$ reduction (Table 9).

\subsection{CXR-ReDonE}

We use three main metrics to evaluate CXR-ReDonE. Each are semantic-based metrics, given that general NLG metrics like BLEU or CIDEr are inadequate for radiology report evaluation due to their poor performance in judging factual accuracy and consistency \citep{brown2020language}. First, we apply \textsc{BERTScore}, which uses contextual embeddings to measure the semantic similarity between the ground truth report and the generated report. We also employ $s_{emb}$, which calculates the cosine similarity between the final hidden representations of both the generated and ground truth reports when passed through a CheXbert labeler \citep{endo2021retrieval, smit2020chexbert}. Finally, we consider RadGraph $F_1$, a metric proposed by \cite{Yu2022.08.30.22279318} that makes use of a RadGraph model to evaluate the overlap in clinical entities included in both the generated and ground truth reports \citep{jain2021radgraph}. 

We find that training CXR-ReDonE on MIMIC-PRO leads to appreciable performance increases, with broad improvements on our expert-edited test set (Section \ref{sec:33}) for both report-level retrieval and sentence-level retrieval with $k\in \{1,2,3\}$, on a variety of evaluation metrics (Table 4). Specifically, for $k=1$, CXR-ReDonE attains an $s_{emb}$ value of 0.3967 ($\Delta + 2.24\%$ over the highest performing baseline). For $k=2$, CXR-ReDonE achieves a \textsc{BERTScore} of 0.2351 ($\Delta + 2.57\%$). Finally, $k=3$ yields CXR-ReDonE's highest RadGraph $F_1$ score of 0.1112 ($\Delta + 2.68\%$). 

We also qualitatively compare the generated reports across training sets (Table 5). Our method does particularly well in generating clinically factual text without hallucinatory references to priors (in the given example, CXR-ReDonE achieves \textsc{BERTScore} $=0.228$, $s_{emb}=0.790$). While the reports generated by CXR-ReDonE contain different verbiage from that of the ground truth, they are still diagnostically accurate and---most importantly---contain far fewer references to priors than those of the baseline CXR-RePaiR model.

\section{Conclusion}

Automatic report generation systems have considerable potential to streamline radiological pipelines but continue to suffer from hallucinations to non-existent prior reports. FilBERT+GPT-3 and GILBERT both demonstrate a generalized capability to broadly improve the factual completeness and consistency of report generation models by removing prior references within radiology report training corpora. As proof of this approach's efficacy, we find that CXR-ReDonE (trained on the GILBERT-edited MIMIC-PRO dataset) produces radiology reports with significantly fewer hallucinatory references to priors. Altogether, these methods show promise in improving the automatic generation of clinically accurate and actionable radiology reports.


\section*{Acknowledgments}

The authors would like to thank Dr. Kibo Yoon, Patricia S. Pile, and Pia G. Alfonso for their central role in developing the MIMIC-PRO test set with prior references manually removed or replaced with clinically accurate statements. The authors would also like to thank Jaehwan Jeong for his advice regarding the technical implementation of CXR-ReDonE, as well as Ethan Chi for advice on the development of FilBERT and GILBERT.

\bibliographystyle{unsrtnat}
\bibliography{references}


\appendix
\section{Training Details}

\begin{table*}[!ht]
    \small
    \centering
    \begin{tabular}{lll}
    \toprule
    \textbf{Temperature} & \textbf{${n}$}  & \textbf{$F_1$} \\
    \midrule
    \textbf{0.3} & \textbf{1} & \textbf{0.5569}     \\ 
    0.4 & 1 & 0.55526     \\ 
    0.2 & 1 & 0.55196     \\ 
    0.0 & 1 & 0.55187     \\ 
    0.1 & 1 & 0.55098    \\ 
    0.3 & 2 & 0.47102     \\ 
    0.3 & 4 & 0.44861     \\ 
    0.3 & 3 & 0.44684 \\
    \bottomrule
    \end{tabular}
    \caption{Hyperparameter search for GPT-3 DaVinci. ${n}$ denotes the number of sentences included in the subreport. Rows are sorted in decreasing order of $F_1$ score.}
   
\end{table*}

\begin{table*}[!ht]
    \small
    \centering
    \begin{tabular}{ll}
    \toprule
    \textbf{Model}  & \textbf{$F_1$}  \\
    \midrule
    \textbf{DaVinci}  & \textbf{0.5569}     \\ 
    Curie  & 0.37263 \\
    Ada & 0.32522 \\
    Babbage & 0.23380 \\

    \bottomrule
    \end{tabular}
    \caption{Performance of all GPT-3 models. All variables besides model type are held constant (temperature = 0.3, $n$ = 1).}
\end{table*}

\subsection{FilBERT}
We develop FilBERT using PyTorch and Tensorflow.  In particular, we load in the BioBERT weights available under the name \texttt{dmis-lab/biobert-base-cased-v1.2} in Huggingface’s \texttt{transformers} library, then finetune solely the sequence classification head. Additionally, we tokenize our data using the BioBERT-specific tokenizer.

We train FilBERT for 10 epochs, using the Adam optimization algorithm. We use a batch size of 16; we also set the learning rate to 2e-5 and $\epsilon$ = 1e-8. We train FilBERT using a single Tesla P100-PCIE-16GB GPU.

\subsection{GILBERT}
Using PyTorch, we train GILBERT for 10 epochs, each with 100 steps, using a training batch size of 4 and a test batch size of 2. We use a gradient clip norm of 10 and an Adam optimizer with learning rate 1e-5. We train GILBERT using a single Tesla T4 GPU.

\subsection{CXR-ReDonE}

With the rephrased reports from MIMIC-PRO, we train CXR-ReDonE over a period of 60 epochs, each with 100 steps. We complete the training cycle using 4 Quadro RTX-8000 GPUs.

\section{Additional Experiments with GPT-3}

\subsection{GPT-3 Hyperparameter Search: Large Language Models are Sentence-Level Learners}\label{apd:first} 
In our few-shot learning approach, we provide context examples to GPT-3 in the prompt. We frame the problem as a text generation task, where the rewritten report (denoted \textit{Edited medical report to remove references to prior medical reports}) is created based on the original report (denoted \textit{Original medical report}).

We empirically determine that a temperature of 0.3 yields the highest accuracy. Additionally, we find that labeling an entire radiology report as an \textit{Original medical report} lowers performance. Therefore, we split each report in the prompt into subreports of length $n \in\{1,2,3,4\}$ sentences and investigate the relationship between $n$ and GPT-3 performance. We find that GPT-3 learns best when $n=1$, that is, when each sentence in the report is fed in as its own subreport (Table 6).

\subsection{Alternative GPT-3 Models Perform Poorly}
We find that alternative GPT-3 variants (Curie, Babbage, and Ada) each perform worse than DaVinci, given their smaller size and lower complexity (Table 7).




\section{Ungrammatical Generations}
GILBERT can sometimes generate ungrammatical sentences---most notably, verbless phrases such as “The cardiomediastinal and hilar contours.” This is a consequence of GILBERT directly \textit{removing} references to priors, rather than \textit{rewriting} them to adjust for grammatical concerns. As a result, MIMIC-PRO  contains a limited number of ungrammatical sentences. Future work could investigate adding a final Transformer-based grammatical error correction layer to CXR-ReDonE to ensure that reports remain grammatical while removing hallucinated references to priors \citep{omelianchuk2020gector}.


\section{Supplementary Tables}\label{apd:sec}

\begin{table*}[h]
    \small
    \centering
    \begin{tabular}{ll}
    \toprule
    \textbf{Sentence} & \textbf{Prior} \\
    \midrule
    No significant interval \textbf{change}. & Yes \\
    \midrule
    No evidence of active \textbf{changes} & Yes \\
    from chronic tuberculosis infection. \\
    \midrule
    Emphysematous \textbf{changes} are    & No \\
    identified.\\
    \midrule
    Midfoot degenerative \textbf{changes}.   & No \\
    \midrule
    There are atherosclerotic \textbf{changes} & No \\
    of the aorta. \\
    \midrule
    Arthritic \textbf{changes} of the spine  & No \\
    are present.\\
    \midrule
    Bony \textbf{changes} of renal osteodystrophy & No \\
    are noted. \\
    \midrule
    Degenerative \textbf{changes} in the spine. & No \\

    \bottomrule
    \end{tabular}
    \caption{As stated in the main body of the paper, the keyword \textit{change} sometimes does not refer to prior reports, due to a diversity of possible usages. Here, we outline sample usages of \textit{change} that do and do not serve as references to priors.}
\end{table*}

\begin{table*}[!ht]
%
\centering
    \begin{tabular}{lll}
    \toprule
    \textbf{Keyword} & \textbf{MIMIC-CXR} & \textbf{MIMIC-PRO} \\
    \midrule
    \textit{Total} & \textit{259376} & \textit{82074}\\
    Change & 58213 & 26403\\
    Unchanged & 30915 & 12956\\
    Prior & 17526 & 3025\\
    Stable & 23837 & 9191\\
    Interval & 15019 & 2379\\
    Previous & 20271 & 2921\\
    Again & 8171 & 247 \\
    Increased & 15019 & 2379\\
    Improve & 18230 & 3153\\
    Remain & 8272 & 3745\\
    Worse & 9651 & 490\\
    Persistent & 12371 & 2596\\
    Removal & 6445 & 6048\\
    Decreased & 5843 & 1768\\
    Similar & 4039 & 826\\
    Earlier & 5082 & 460\\
    Recurrence & 1077 & 645\\
    Redemonstrate & 88 & 0\\
    \bottomrule
    \end{tabular}
    \caption{Number of references to priors, broken down by keyword, in MIMIC-CXR and MIMIC-PRO. It is clear that MIMIC-PRO contains far fewer references to priors than MIMIC-CXR in every category---proof of GILBERT's efficacy.}
   
\end{table*}

\begin{table*}[h!]
\centering

\scriptsize

  {\caption{Additional examples of different methods' reports with removed references to priors compared to the ground truth report. Prior references are color coded to improve readability.\\}}
  {\begin{tabular}{p{0.224\linewidth} p{0.224\linewidth} p{0.224\linewidth} p{0.224\linewidth}}
  \toprule
  \textbf{Original} & \textbf{Ground Truth} & \textbf{FilBERT+GPT-3} & \textbf{GILBERT}\\
  \midrule
  Frontal and lateral radiographs of the chest demonstrate \hlc{Persistent}{persistent} large right perihilar mass, \hlc{Comparison}{which is slightly larger as} \hlc{Comparison}{compared to the \hlc{Prior}{prior} study}. 
  
  & 
  Frontal and lateral radiographs of the chest demonstrate large right perihilar mass. 
  &  
  Frontal and lateral radiographs of the chest demonstrate large right perihilar mass.
  &

  
 Frontal and lateral radiographs of the chest demonstrate large right perihilar mass. \\
  
  This is in a region of \hlc{Prior}{prior} fiducial seed placement, and may correspond to post-radiation changes; however, \hlc{Recurrence}{recurrence} of malignancy cannot be excluded. 
  
  & 
  
  This is in a region of fiducial seed placement, and may correspond to post-radiation changes; however, malignancy cannot be excluded. 
  
  &
  This is in a region of \hlc{Prior}{prior} fiducial seed placement, and may correspond to post-radiation changes; however, malignancy cannot be excluded.
  & 
  This is in a region of fiducial seed placement; however, of malignancy cannot be excluded.  \\

  \hlc{Again}{Again} seen are heterogeneous opacities at the right base, with a small right-sided pleural effusion. 
  
   & 
  
 Seen are heterogeneous opacities at the right base, with a small right-sided pleural effusion. 
  
  &
  Heterogeneous opacities at the right base, with a small right-sided pleural effusion.
  & 
Seen are heterogeneous opacities at the right base, with a small right-sided pleural effusion.   \\

  The left lung 
  is essentially clear. \hlc{Unchanged}{The cardiomediasti}
  \hlc{Unchanged}{nal and hilar contours are}
  \hlc{Unchanged}{unchanged.} There is no pneumothorax or focal consolidation. 
  
  & 
  
The left lung is essentially clear.  There is no pneumothorax or focal consolidation.
  
  &
  The left lung is essentially clear. Cardiomediastinal and hilar contours. There is no pneumothorax or focal consolidation.
  & 
 The left lung is essentially clear.  The cardiomediastinal and hilar contours. There is no pneumothorax or focal consolidation.  

  \\
Right lung opacities have \hlc{Worse}{slightly worsened} \hlc{Worse}{since previous exam} and are \hlc{Comparison}{slightly more} confluent, suspicious for an infectious process or aspiration. 
    
  & 
  
Right lung opacities are confluent, suspicious for an infectious process or aspiration. 

  &
  
Right lung opacities are \hlc{Comparison}{slightly more} confluent, suspicious for an infectious process or aspiration.

& 
Right lung opacities are confluent, suspicious for an infectious process or aspiration. 
\\
No acute cardiopulmonary process. \hlc{Change}{No} \hlc{Change}{significant interval change.} 
&
No acute cardiopulmonary process.
&
No acute cardiopulmonary process. \hlc{Change}{No} \hlc{Change}{significant interval change.} 
&
No acute cardiopulmonary process.
\\

  \bottomrule
\end{tabular}}

  {\begin{tabular}{c c c c c c c c c c}
  \hlc{Persistent}{persistent} &  \hlc{Comparison}{comparison} & \hlc{Prior}{prior} & \hlc{Recurrence}{recurrence} & \hlc{Again}{again} &
  \hlc{Worse}{worse} & 
  \hlc{Change}{change} &
  \hlc{Unchanged}{unchanged}\\
  \end{tabular}}
\end{table*}

\begin{table*}[!ht]
\centering

\scriptsize

  {\caption{Additional examples of different methods' generated reports compared to the expert-edited ground truth report.\\}}
  {\begin{tabular}{p{0.06\linewidth}p{0.28\linewidth}p{0.28\linewidth}p{0.28\linewidth}}
  \toprule
  \textbf{Type} & \textbf{Ground Truth} & \textbf{MIMIC-CXR Model} & \textbf{MIMIC-PRO Model}\\
  \midrule

Report-level & AP chest: There is substantial engorgement and indistinctness of pulmonary vessels, consistent with the clinical impression of pulmonary edema. There is some indistinctness and engorgement of pulmonary vessels, consistent with the clinical impression of elevated pulmonary venous pressure. 

 &
Heart size and mediastinum \hlc{Stable}{stable}. Minimal interstitial opacities in the lung bases are but it might be related test technique of the chest radiograph. There is no definitive consolidation to suggest infection but attention to the lung bases is recommended. There is no pneumothorax. There is no pleural effusion.

&

Heart size and mediastinum are overall unremarkable except for mild vascular congestion and minimal interstitial edema. There is no appreciable pleural effusion. There is no pneumothorax. Lung volumes are low.
\\
$k=1$
& 
AP chest: Extensive opacification in both lungs, sparing the left mid and lower lung zone. AP chest: Severe widespread pulmonary infiltration is significant, with near confluence of opacification in the left lung, and moderate-to-severe left pleural effusion.

&
Widespread parenchymal consolidations involving the left lung as well as loculated pleural effusion \hlc{Unchanged}{unchanged.}

&
Diffuse bilateral airspace process suggestion of areas of cavitation particularly on the left side and a confluent opacification in the right mid upper lung.

\\

$k=2$ & PA and lateral chest:  There is extensive consolidation, predominantly at the base of the right lung in the lower lobe, probably also in the right middle lobe and to a small extent in the mid left lung.

&
Large scale right lower lobe pneumonia \hlc{Unchanged}{unchanged}. Right pleural effusion and right lower lung consolidation for infectious process.
&
Right upper and lower lobe airspace consolidation obscuring a known right hilar mass may represent hemorrhage or pneumonia. Extensive consolidation.
\\

$k=3$ & The opacification at the right base appears to be significant, raising the possibility of superimposed pneumonia in this patient with hyperexpansion of the lungs consistent with chronic pulmonary disease. There is opacity at right greater than left lung apices. 
 &
Asymmetric opacity in the right infrahilar region could reflect an early bronchopneumonia versus atelectasis. Subtle asymmetric \hlc{Increased}{increased} opacity in the right lower lobe could reflect concurrent pneumonia or aspiration in the appropriate clinical situation, however \_\_\_. Opacity in the right infrahilar region may reflect early bronchopneumonia.
&
Streaky right basilar opacity may be atelectasis although aspiration or infection would be difficult to exclude. Opacity in the right lung base, which could represent atelectasis but cannot exclude pneumonia or aspiration in the right clinical setting. Peribronchial infiltration right lower lobe could be chronic scarring with atelectasis and possible bronchiectasis, but would need ct evaluation for anatomic diagnosis and to establish chronicity.

\\

 


  \bottomrule
\end{tabular}}

  {\begin{tabular}{c c c}
  \hlc{Stable}{stable}  & \hlc{Increased}{increased} & \hlc{Unchanged}{unchanged}\\
  \end{tabular}}\\

  
\end{table*}

\end{multicols}
\end{document}